\documentclass{article}

\usepackage{microtype}
\usepackage{graphicx}
\usepackage{subfigure}
\usepackage{booktabs} % for professional tables

\usepackage{hyperref}

\def\mB{{\mathcal B}}

\def\mD{{\mathcal D}}

\DeclareMathAlphabet\mathbfcal{OMS}{cmsy}{b}{n}
\def\0{{\bf 0}}
\def\1{{\bf 1}}

\def\bb{{\bf b}}

\def\bb{{\bf b}}

\newtheorem{remark}{Remark}

\def\eg{\emph{e.g.}} 
\def\ie{\emph{i.e.}}

\def\wrt{{w.r.t.~}}

\usepackage{amsmath}

\def\sss{\scriptscriptstyle} 

\usepackage[accepted]{icml2021}

\usepackage{times}
\usepackage{epsfig}
\usepackage{graphicx}
\usepackage{amsmath}
\usepackage{amssymb}

\usepackage{booktabs}       % professional-quality tables
\usepackage{amsfonts}       % blackboard math symbols
\usepackage{nicefrac}       % compact symbols for 1/2, etc.
\usepackage{microtype}      % microtypography
\usepackage{multirow}
\usepackage{xspace}
\usepackage{enumitem}
\usepackage{relsize}
\usepackage{subfigure}

\usepackage{amsmath}
\usepackage{bbm}
\usepackage{wrapfig}

\usepackage{amssymb}% http://ctan.org/pkg/amssymb
\usepackage{pifont}% http://ctan.org/pkg/pifont

\def\mytitle{\kui{Pareto-Frontier-aware Neural Architecture Generation for Diverse Budgets}}
\newcommand{\eat}[1]{}

\newcommand{\sexyname}{NAG\xspace}

\def\lzp{\textcolor{cyan}}
\def\guo{\textcolor{blue}}

\def\kui{\textcolor{red}}
\def\qi{\textcolor{magenta}}
\def\lhk{\textcolor[RGB]{0, 139, 0}}

\def\lzp{\textcolor{black}}
\def\guo{\textcolor{black}}

\def\qi{\textcolor{black}}
\def\lhk{\textcolor{black}}
\def\kui{\textcolor{black}}

\icmltitlerunning{Pareto-Frontier-aware Neural Architecture Generation for Diverse Budgets}

\begin{document}

\twocolumn[
\icmltitle{\mytitle}

% It is OKAY to include author information, even for blind
% submissions: the style file will automatically remove it for you
% unless you've provided the [accepted] option to the icml2021
% package.

% List of affiliations: The first argument should be a (short)
% identifier you will use later to specify author affiliations
% Academic affiliations should list Department, University, City, Region, Country
% Industry affiliations should list Company, City, Region, Country

% You can specify symbols, otherwise they are numbered in order.
% Ideally, you should not use this facility. Affiliations will be numbered
% in order of appearance and this is the preferred way.
\icmlsetsymbol{equal}{*}

\begin{icmlauthorlist}
\icmlauthor{Yong Guo}{scut}
\icmlauthor{Yaofo Chen}{scut}
\icmlauthor{Yin Zheng}{weixin}
\icmlauthor{Qi Chen}{scut}
\icmlauthor{Peilin Zhao}{ailab}
\icmlauthor{Jian Chen}{scut}
\icmlauthor{Junzhou Huang}{ailab}
\icmlauthor{Mingkui Tan}{scut}

\end{icmlauthorlist}

\icmlaffiliation{scut}{School of Software Engineering, South China University of Technology}
\icmlaffiliation{weixin}{Weixin Group, Tencent}
\icmlaffiliation{ailab}{Tencent AI Lab, Tencent}
% \icmlaffiliation{uta}{University of Texas at Arlington}
%\icmlaffiliation{klab}{Guangdong Key Laboratory of Big Data Analysis and Processing}
%\icmlaffiliation{pzlab}{Pazhou Laboratory}

\icmlcorrespondingauthor{Mingkui Tan}{mingkuitan@scut.edu.cn}
\icmlcorrespondingauthor{Jian Chen}{ellachen@scut.edu.cn}

% You may provide any keywords that you
% find helpful for describing your paper; these are used to populate
% the "keywords" metadata in the PDF but will not be shown in the document
\icmlkeywords{Machine Learning, ICML}

\vskip 0.3in
]

% this must go after the closing bracket ] following \twocolumn[ ...

% This command actually creates the footnote in the first column
% listing the affiliations and the copyright notice.
% The command takes one argument, which is text to display at the start of the footnote.
% The \icmlEqualContribution command is standard text for equal contribution.
% Remove it (just {}) if you do not need this facility.

\printAffiliationsAndNotice{}  % leave blank if no need to mention equal contribution
%\printAffiliationsAndNotice{\icmlEqualContribution} % otherwise use the standard text.

\begin{abstract}
Designing feasible and effective architectures under diverse computation budgets incurred by different applications/devices is essential for deploying deep models in practice.
Existing methods often perform an independent architecture search for each target budget, which is very inefficient yet unnecessary.
Moreover, the repeated independent search manner would inevitably ignore the common knowledge among different search processes and hamper the search performance.
To address these issues, we seek to train a general architecture generator that automatically produces effective architectures for an arbitrary budget merely via model inference.
To this end, we propose a Pareto-Frontier-aware Neural Architecture Generator (\sexyname) which takes an arbitrary budget as input and produces the Pareto optimal architecture for the target budget.
We train \sexyname by learning the Pareto frontier (\ie, the set of Pareto optimal architectures) over model performance and computational cost (\eg, latency).
Extensive experiments on three platforms (\ie, mobile, CPU, and GPU) show the superiority of the proposed method over existing NAS methods.
\end{abstract}

\section{Introduction}

Deep neural networks (DNNs)~\cite{lecun1989backpropagation} have been the workhorse of many challenging tasks, including image classification~\cite{krizhevsky2012imagenet} and 
semantic segmentation~\cite{long2015fully}. 
% and object detection~\cite{zhao2019object,tian2019fcos}.
However, designing effective architectures often relies heavily on human expertise.
To alleviate this, neural architecture search (NAS) has been proposed to automatically design architectures~\cite{zoph2016neural}. Existing studies show that the automatically searched architectures often outperform the manually designed ones~\cite{liu2018darts,liu2017hierarchical}. 

However, deep models often contain a large number of parameters and come with \lhk{a} high computational cost.
As a result, it is hard to deploy deep models to real-world scenarios with limited computation resources.
{Thus, we have to carefully design architectures to fulfill a specific budget.}
More critically, we may have different budgets of computation resources in \lhk{the} real world.
For example, a company may develop/maintain multiple applications and each of them has a specific budget of latency. 

\qi{Existing methods~\cite{tan2019mnasnet,stamoulis2019single} seek to design a model to obtain an architecture under a single budget. When we consider diverse budgets, they have to conduct multiple search processes for each budget~\cite{tan2019mnasnet},} which is very inefficient yet unnecessary.
Besides finding a single architecture, 
one can learn a generator model that produces architectures according to a learnable distribution~\cite{xie2019exploring,ru2020neural}. However, the produced architectures tend to have a low variance in terms of model performance and computational cost~\cite{xie2019exploring,ru2020neural}. 
As a result, these methods still have to learn a generator for each budget.
Unlike these methods, one can also exploit the population-based methods to simultaneously find multiple architectures and then select an appropriate one from them to fulfill a specific budget~\cite{lu2019nsga,lu2020nsganetv2}. However, due to the limited population size, the selected architecture does not necessarily satisfy the considered budget. Thus, how to design architectures under diverse budgets remains an open question.

\begin{figure*}
    \centering
	\subfigure[Diverse application scenarios with different budgets.]{
		\includegraphics[width = 1.19\columnwidth, height = 0.53\columnwidth]{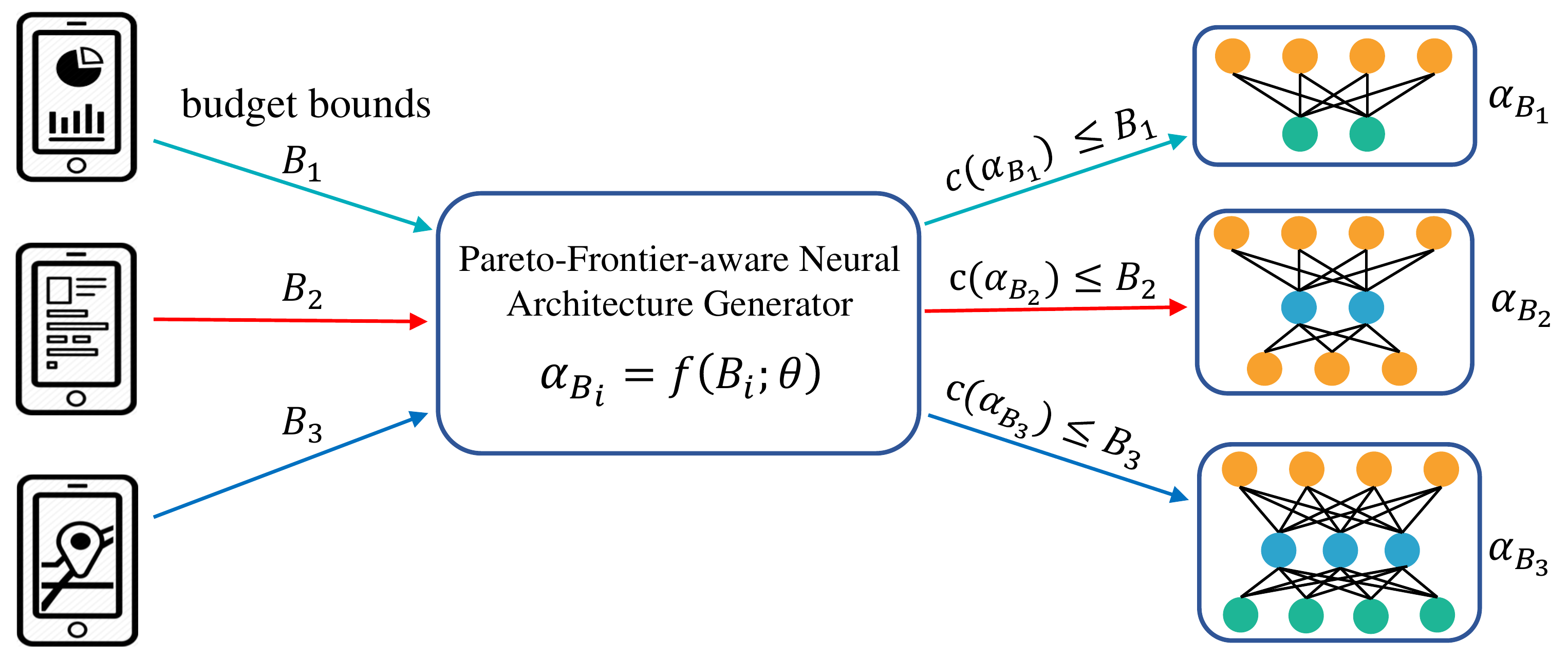}\label{fig:application_sub}
	}~
	\subfigure[\kui{Comparisons of NAG and conventional NAS methods.}]{
		\includegraphics[width = 0.85\columnwidth]{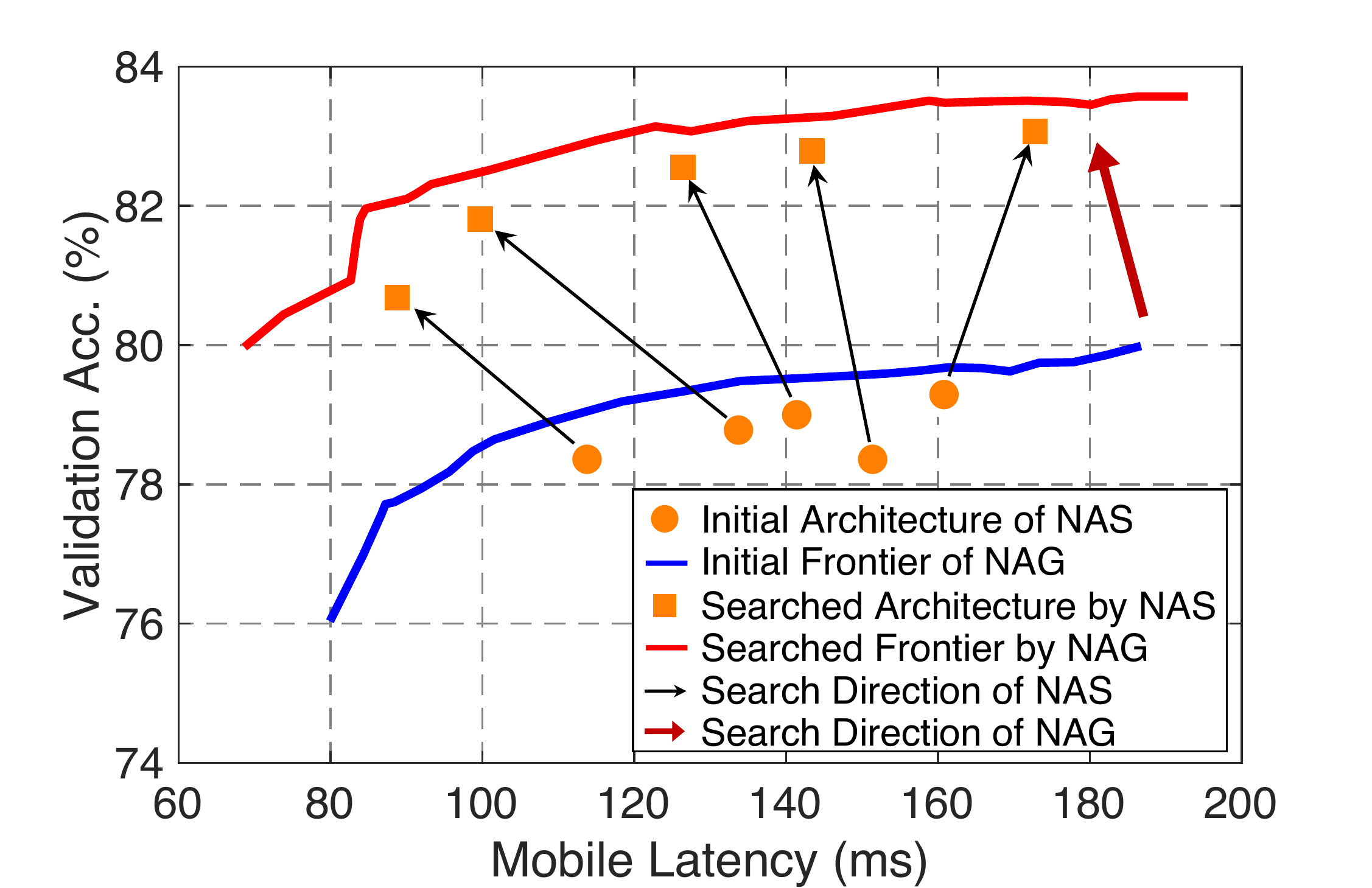}\label{fig:search_direction}
	}
    \caption{An illustration of the applications of \sexyname and the comparisons of search strategies. (a) \sexyname takes {an} arbitrary budget as input and output{s} the corresponding architecture.
    (b) \sexyname learns the Pareto frontier rather than optimizing a specific architecture.
    }
    \label{fig:application}
    % \vspace{-13 pt}
\end{figure*}

In this paper, rather than finding an architecture for a specific budget, we seek to learn an architecture generator to produce feasible architectures under diverse budgets.
Note that we consider the same search space when designing architectures for different budgets.
One may find some common architectures that perform well across multiple budgets.
For example, based on a common architecture under some budgets, we may easily obtain promising architectures for a larger/smaller budget by slightly modifying some of its elements/layers.
In this sense, it is possible to find effective architectures for diverse budgets simultaneously.
More critically, the knowledge provided by these common architectures may also benefit the search process (See results in Table~\ref{tab:mobile_comp}).
% \qi{In this way, the model optimized under diverse budgets shares a common search space, which is able to facilitate the efficiency and performance of the searched architectures.}
Inspired by this, we propose a Pareto-Frontier-aware Neural Architecture Generator (\sexyname) 
which takes a budget as input and produces \lhk{architectures} satisfying the considered budget (See Figure~\ref{fig:application_sub}). 
% Since the NAS problems under different budgets share the same search space, 
Note that the Pareto optimal architectures under different budgets should lie on a distribution, \ie, the Pareto frontier over model performance and computational cost~\cite{kim2005adaptive}.
We propose to learn the Pareto frontier (\ie, improving the blue curve to the red curve in Figure~\ref{fig:search_direction}) instead of finding multiple discrete Pareto optimal architectures independently.
To achieve this, we evenly sample a set of discrete budgets from the range of possible values and maximize the expected reward of the searched architectures over these budgets to approximate the Pareto frontier.
To evaluate architectures under diverse budgets, we design an architecture evaluator to learn a Pareto dominance rule which determines whether an architecture is better than another. 
During inference, given an arbitrary budget, we are able to produce promising architectures merely via model inference, which is very efficient.

\begin{figure*}
    \centering
    \includegraphics[width=0.87\textwidth]{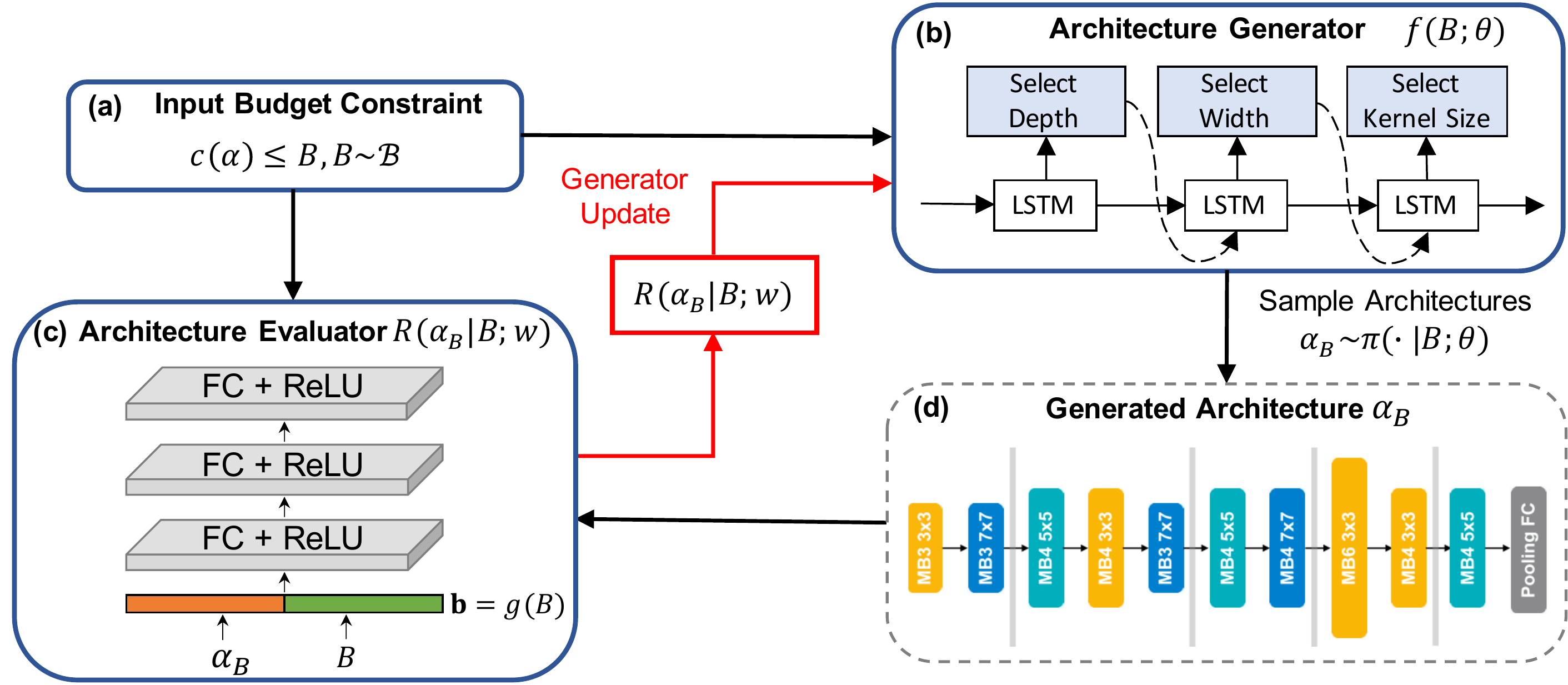}
    \caption{Overview of the proposed \sexyname. 
    % Our \sexyname takes a budget constraint as input and produces promising architectures that satisfy the budget constraints. 
    \qi{Our \sexyname mainly consists of two modules: an architecture generator $f(\cdot;\theta)$ and an architecture evaluator $R(\cdot|\cdot;w)$. Specifically, we build the generator model based on an LSTM network, which takes a budget constraint $B$ as input and produces a promising architecture $\alpha_B$ that satisfies the budget constraint, \ie, $c(\alpha)$. To optimize the generator model, we design the evaluator using three fully connected (FC) layers to estimate the performance of the generated architectures $\alpha_B$. The orange and green boxes in (c) denote the embeddings of architecture $\alpha_{\sss B}$ and the budget w.r.t. $B$, respectively. }}
    \label{fig:overview}
\end{figure*}

We summarize the contributions of our paper as follows.
\begin{itemize}
    % \item We propose a Pareto-Frontier-aware Neural Architecture Generator (\sexyname) model that {produces effective architectures for any given computation budget.}
    % Unlike NAS that optimizes a specific architecture, we learn the whole Pareto frontier over model performance and computational cost (\eg, latency). 
    \item 
    \guo{
    Rather than designing architectures for a specific budget, 
    we consider learning to generate optimal architectures for any given budget.} To this end, 
    we propose a Pareto-Frontier-aware Neural Architecture Generator (\sexyname) method to learn the Pareto frontier of architectures over model performance and computational cost
    % rather than optimizing a specific architecture 
    (as shown in Figure~\ref{fig:overview}).
    % that takes an budget as input to produce feasible architectures.
    % \qi{To this end, we seek to learn the Pareto frontier over model performance and computational cost (As shown in Figure~\ref{fig:overview}).}

    % % By taking such a rule as the reward, our \sexyname is able to iteratively find better frontiers to approach the ground-truth Pareto frontier.
    % \item \qi{Based on the architecture evaluator,} we propose a Pareto-Frontier-aware Neural Architecture Generator (\sexyname) that produces effective architectures for any given computation budget.
    % \qi{To this end, we seek to learn the Pareto frontier over model performance and computational cost (As shown in Figure~\ref{fig:overview}).}
    % % Unlike NAS that optimizes a specific architecture, we learn the whole Pareto frontier over model performance and computational cost (\eg, latency). 
    
    \item 
    To learn the generator, we propose an architecture evaluator to evaluate architectures under diverse budgets.
    Moreover, to make the Pareto frontier learning feasible,
    % To learn the architecture generator to produce the Pareto frontier,  we build an architecture evaluator to evaluate architectures under diverse budgets.
    % To learn the architecture evaluator,
    we propose a Pareto dominance rule to determine whether an architecture is better than another. 
    % Based on the Pareto dominance rule, we learn an evaluator to estimate performance of architectures.

    \item Extensive experiments on three hardware platforms show that the architectures produced by our \sexyname consistently outperform the architectures searched by existing methods under different budgets. 
\end{itemize}

\section{Related Work}

% In this section, we first revisit the standard neural architecture search (NAS) methods. Second, we discuss the methods to design architectures under resource constraints. Third, we discuss the methods to learn the Pareto frontier.

\textbf{Neural Architecture Search (NAS)}. \kui{Unlike manually designing architectures with expert knowledge, NAS seeks to automatically design more effective architectures.}
\lzp{However, solving the NAS problem is challenging due to the non-convex nature and the large search space~\cite{pham2018efficient,guo2020breaking}.}
\lhk{Existing NAS methods 
% use different search stragetes to address this problem, which 
can be roughly divided into three categories, namely, reinforcement-learning-based methods, evolutionary approaches, and gradient-based methods.}
\lzp{Specifically, reinforcement-learning-based methods~\cite{zoph2016neural, pham2018efficient, tan2019mnasnet} learn a controller to produce architectures. Evolutionary approaches~\cite{real2017large, liu2017hierarchical, real2019regularized} search for promising architectures by gradually evolving a population. 
% Different from these methods that search within a discrete space, 
Gradient-based methods~\cite{liu2018darts, chen2019progressive, xu2020pcdarts} relax the search space to be continuous and optimize architectures by gradient descent.}
% \citet{zoph2016neural} use reinforcement learning to discover the optimal configuration of each layer.
% % AmoebaNet-A~\cite{real2019regularized} employs evolution algorithms and proposes a new regularization method. 
% \citet{liu2018darts} present a differentiable NAS method by relaxing the search space to be continuous. 
% % PNAS~\cite{liu2018progressive} gradually enlarges the search space and performs architecture search in a progressive manner.
% % However, these methods ignore the resource constraints of real-world applications when performing architecture search.
Unlike these methods \lhk{that find a single architecture}, one can design \lhk{different} architectures by training an architecture generator. RandWire~\cite{xie2019exploring} designs stochastic network generators to generate randomly wired architectures.
% , which achieve competitive performance with NAS methods.
% To further optimize the network generators, NAGO~\cite{ru2020neural} proposes a newly designed hierarchical and graph-based search space.
NAGO~\cite{ru2020neural} proposes a hierarchical and graph-based search space to reduce the optimization difficulty.
% NAGO~\cite{ru2020neural} further optimizes the network generators in a newly designed hierarchical and graph-based search space. 
However, these generated architectures tend to have a low variance of both model performance and computational cost~\cite{xie2019exploring,ru2020neural}.
Besides, given diverse budgets, they also have to learn a generator model for each budget to produce feasible architectures.

% \vspace{3pt}
\textbf{Architecture Design under Resource Constraints}. \lhk{Many efforts have been made in designing architectures under a resource constraint~\cite{cai2019once, huang2020ponas, elsken2018efficient, Bender2020TuNAS, guo2020single}.}
\lhk{To reduce training budget,} OFA~\cite{cai2019once} trains a powerful supernet, from which we can directly get a specialized sub-network without additional training.
% \lhk{To satisfy the constraints,}
Moreover, PONAS~\cite{huang2020ponas} builds an accuracy table to find architectures satisfying a single budget constraint.
{TuNAS~\cite{Bender2020TuNAS} proposes a reward function to restrict the latency of the searched architecture, which omits additional hyper-parameter tuning.}
However, given various computation budgets, these methods need to repeat the architecture search process for each budget. By contrast, our \sexyname only needs to search once to produce architectures that satisfy arbitrary budgets.
% \cyf{Add discussion of TuNAS!!!} 

\textbf{Pareto Frontier Learning.}
Given multiple objectives, Pareto frontier learning aims to find a set of Pareto optimal solutions over them.
% Most methods convert the problem into a single-objective problem by constructing a weighted sum/product utility function~\cite{wierzbicki1982mathematical,miettinen2012nonlinear}.
% To simultaneously find multiple Pareto optimal solutions (\ie, Pareto frontier), 
Most methods exploit evolutionary algorithms~\cite{deb2002fast,kim2004spea} to solve this problem.
\lhk{Inspired by them, many efforts have been made to simultaneously find a set of Pareto optimal architectures over accuracy and computational cost~\cite{cheng2018searching,dong2018dpp}.}
% Recently, some NAS methods have been proposed to simultaneously find a set of Pareto optimal architectures over accuracy and computational cost~\cite{cheng2018searching,dong2018dpp}.
% choose an appropriate solution on the trade-off frontier
% for implementation???
% NSGA-Net~\cite{lu2019nsga} proposes an evolutionary approach to find a set of trade-off architectures over multiple objectives in a single run. 
% \lhk{To find a set of trade-off architectures over multiple objectives in a single run, NSGANetV1~\cite{lu2020multi} 
% proposes an evolutionary approach.}
Recently, NSGANetV1~\cite{lu2020multi} presents an evolutionary approach to find a set of trade-off architectures over multiple objectives in a single run.
% exploits a set of architectures to approximate the Pareto frontier by recombining and
% modifying architectural components progressively.
NSGANetV2~\cite{lu2020nsganetv2} \lhk{further} presents two surrogates (at the architecture and weights level) to produce task-specific models under multiple competing objectives.
% However, it is still unknown how to learn the Pareto frontier in NAS.
Given a target budget, these methods may manually select an appropriate architecture from \lhk{a} set of searched architectures.
However, given limited population size, the selected architectures do not necessarily satisfy the considered budget.
More critically, due to the limited population size, these methods may only find a small number of architectures on the Pareto frontier.
% these methods only approximate the Pareto frontier by some discrete solutions.
Thus, how to learn the entire Pareto frontier of architectures still remains an open question.

\section{Pareto Neural Architecture Generation}\label{sec:method}

% In this paper, we propose a Pareto-Frontier-aware Neural Architecture Generator (\sexyname) model that automatically produces effective architectures for any given budget. 
% We show the overview of \sexyname in Figure~\ref{fig:overview}.

% \subsection{Problem Definition}
\noindent \textbf{Notation.} 
\kui{In this paper, we \lhk{consider producing} architectures under diverse budgets simultaneously. For convenience, let $B$ be a computation budget (\eg, latency and MAdds) which can be considered as a random variable drawn from some distribution $\mB$, namely $B {\sim} \mB$.}
% , such as latency, the number of multiply-adds (MAdds) and memory consumption.
\qi{
Let $\Omega$ be an architecture search space. For any architecture $\alpha \in \Omega$, we use $c(\alpha)$ and ${\rm Acc}(\alpha)$ to measure the cost and validation accuracy of $\alpha$, respectively.}
% We evaluate $\alpha$ using a reward function $R(\alpha; w)$ parameterized by $w$.
Without loss of generality, 
we use $\alpha_{\sss B}^{\sss (i)}$ to denote the $i$-th generated architecture under the budget constraint $c(\alpha_{\sss B}^{\sss (i)}) \leq B$.
% We use $\mathbbm{1}[A]$ to denote a function that has $\mathbbm{1}[A]=1$ if $A$ is true and $\mathbbm{1}[A]=-1$ otherwise. 

We focus on the architecture generation problem that aims to produce effective architectures for real-world applications/devices with diverse computation budgets. 
{Note that the optimal architectures under different budgets lie on the Pareto frontier over model performance and computational cost~\cite{kim2005adaptive}. 
Thus, we propose to learn the whole Pareto frontier to enable the generator to produce architectures for the budget with any possible value.}
In this way, we {can} easily find a promising architecture from the learned frontier by restricting the computational cost to be less than the considered budget.

{
% To learn the architecture generator, we cast the optimization problem into a decision making problem and reformulate it as a Markov Decision Process (MDP). 
Formally,
we generate architectures by determining which architecture should be the optimal one \wrt  any given budget.
Specifically, we seek to learn a transformation mapping $B {\rightarrow} \alpha_{\sss B}$ to produce an architecture whose computational cost satisfies the constraint $c(\alpha_{\sss B}) \leq B$.}
Note that 
% we have no supervision signal (\ie, ``ground-truth'' architecture satisfying a budget) and the metrics of 
both accuracy and computational cost in the objective are non-differentiable. We use reinforcement learning~\cite{williams1992simple} to train the generator model (See Algorithm~\ref{alg:training}).

% \vspace{3 pt}
% \noindent {\textbf{MDP formulation details.}}
% A typical MDP is defined by a tuple $(\mathcal{S}, \mathcal{A}, P, {R})$,
% where $\mathcal{S}$ is a finite set of states, $\mathcal{A}$ is a finite set of actions, $P: \mathcal{S} \times \mathcal{A} \times \mathcal{S} \rightarrow \mathbb{R}$ is the state transition distribution, and ${R}: \mathcal{S} \times \mathcal{A} \rightarrow \mathbb{R}$ is the reward function. 
% Here, we define the pair of budget and the corresponding architecture $(T, \alpha)$ as a state, the decision to find a better architecture under the considered budget as an action.
% To obtain the Pareto optimal (also called non-dominated) solutions, we develop a Pareto dominance rule to compute the reward (See Section~\ref{sec:reward}).
% Here, we exploit policy gradient method~\cite{williams1992simple} to solve the MDP problem.

\subsection{Learning the Architecture Generator $f(B;\theta)$} \label{sec:cdnas}

In this section, we propose an architecture generator model {to automatically produce effective architectures for any given computation budget.}
As shown in Figure~\ref{fig:overview}, we develop a conditional model $f(B;\theta)$ \qi{based on an LSTM network. The $f(B;\theta)$} takes a budget $B$ as input and outputs an architecture $\alpha_{\sss B} {=} f(B;\theta)$ satisfying the budget constraint $c(\alpha_{\sss B}) \leq B$. 
Here, $\theta$ denotes the learnable parameters of $f(\cdot)$.
Note that the optimal architecture under a specific budget should lie on the Pareto frontier over model performance and computational cost.
To learn a general generator for {an} arbitrary budget, we seek to learn the Pareto frontier rather than optimizing a specific architecture.

\textbf{NAS under \lhk{a} single budget.}
To illustrate our method, we first revisit the NAS problem with a single budget and then generalize it to the problem with diverse budgets.
Note that it is non-trivial to directly find the optimal architecture~\cite{zoph2016neural}. By contrast, 
% Following~\cite{zoph2016neural}, 
one can first learn a policy $\pi(\cdot; \theta)$ and then conduct sampling from it to find promising architectures, \ie, $\alpha \sim \pi(\cdot; \theta)$. Given a budget $B$, the optimization problem becomes
\begin{equation}\label{eq:obj-single-constraint}
    \begin{aligned}
         \max_{\theta} ~\mathbb{E}_{\alpha \sim \pi(\cdot; \theta)} ~\left[R \left( \alpha|B; w \right)\right], ~\text{s.t. } ~c(\alpha) \leq B.
     \end{aligned}
\end{equation}
Here, $\pi(\cdot;\theta)$ is the learned policy parameterized by $\theta$, and $R(\alpha|B; w)$ is the reward function parameterized by $w$ that measures the joint performance on both the accuracy and the latency of $\alpha$. We use $\mathbb{E}_{\alpha \sim \pi(\cdot; \theta)} \left[  \cdot \right]$ to denote the expectation over the searched architectures. 

\textbf{NAS under multiple budgets.}
Problem~(\ref{eq:obj-single-constraint}) only focuses on one specific budget constraint. In fact, we seek to learn the Pareto frontier over the whole range of budgets (\eg, latency).
However, this problem is hard to solve since there may exist infinite Pareto optimal architectures with different computational cost. To address this, one can learn the approximated Pareto frontier by finding a set of uniformly distributed Pareto optimal points~\cite{grosan2008generating}. In this paper, we evenly sample $K$ budgets from the range of latency and maximize the expected reward over them.
Thus, the optimization problem becomes
\begin{equation}\label{eq:obj-multi-constraint}
    \begin{aligned}
         \max_{\theta} &~\mathbb{E}_{B \sim \mB} \left[  \mathbb{E}_{\alpha_{_B} \sim \pi(\cdot|B; \theta)} ~\left[R \left(\alpha_{\sss B} | B; w \right) \right] \right], \\
         &~\text{s.t. } ~c(\alpha_{\sss B}) \leq B, ~B \sim \mB,
     \end{aligned}
\end{equation}
where 
% $\alpha_T$ denotes the searched architecture under the constraint w.r.t. $T$, 
$\mathbb{E}_{B \sim \mB} \left[  \cdot \right]$ denotes the expectation over the budget. 
Unlike Eqn.~(\ref{eq:obj-single-constraint}), $\pi(\cdot|B;\theta)$ is the learned policy conditioned on the budget of $B$.
To find the architectures satisfying the budget constraint, we take $B$ into account to compute the reward $R(\cdot | B; w)$. 
We will illustrate this in Section~\ref{sec:reward}.

% However, learning the Pareto frontier
% The generator model $f(B;\theta)$ takes any given budget as input and outputs promising architectures to fulfill the budget.
\qi{In practice, we use a policy gradient method to learn the architecture generator. To encourage exploration, we introduce an entropy regularization term $H(\cdot)$ to measure the entropy of the policy.}
% We learn the generator with policy gradient and use an entropy regularization term to encourage exploration.
Thus, the objective becomes
\begin{equation}\label{eq:obj_entropy}
\small
J(\theta) {=} \mathbb{E}_{B {\sim} \mB} \left[ \mathbb{E}_{\alpha_{_B} {\sim} \pi(\cdot |B;\theta)}  \left[ {R}\left(\alpha_{\sss B} | B; w \right) \right]  {+} \lambda H \big(\pi(\cdot|B; \theta) \big)  \right],
\end{equation}
where 
% $H(\cdot)$ evaluates the entropy of the policy and
$\lambda$ is a hyper-parameter.
In each iteration, we first sample $\{B_k\}_{k=1}^{K}$ from the distribution $\mB$, and then sample $N$ architectures $\{\alpha_{\sss {B_k}}^{\sss (i)}\}_{i=1}^{N}$ for each budget $B_k$.
Thus, the gradient of the objective for the generator w.r.t. $\theta$ becomes\footnote{We put the derivations of the gradient in the supplementary.}
\begin{equation} \label{eq:entropy_gradient}
\small
    \begin{aligned}
    \nabla_{\theta} J(\theta) {\approx} &\frac{1}{KN} \sum_{k=1}^{K} \sum_{i=1}^{N} \Big[ \nabla_{\theta} \log \pi(\alpha_{\sss {B_k}}^{(i)}|B_k;\theta) 
    % \textcolor{blue}
    {R(\alpha_{\sss {B_k}}^{(i)}|B_k;w)} \\
    &+ \lambda \nabla_{\theta} H(\pi(\cdot|B_k;\theta)) \Big].
    \end{aligned}
\end{equation}

% Vector Representation 
% evaluator

\qi{
To learn the architecture generator, 
% there still remain two problems: 
\guo{we still have to answer two questions.}
1) How to represent the budget bound $B$ as the inputs of \sexyname? 
As mentioned before, our \sexyname consider $K$ discrete budgets during training. 
% The value of budget may vary a lot from case to case (\eg, 10ms and 1000ms) and may hamper the training of \sexyname. 
% To alleviate this issue, 
To represent different budgets, 
we use an embedding vector~\cite{pham2018efficient} to represent different budgets (See details in Section~\ref{sec:rep_budget}). 
2) How to define the reward function $R\left(\cdot| B; w \right)$ in Eqn.~(\ref{eq:entropy_gradient}) to evaluate the generated architectures?
Given diverse budgets, an architecture should be assigned a specific reward score for each budget. However, it is non-trivial to manually design the reward function. Instead, we propose to learn an architecture evaluator to automatically predict the score (See details in Section~\ref{sec:reward}).
% We will depict the details of the budget representation method and the method to obtain $R\left(\cdot| B; w \right)$ in Sections~\ref{sec:rep_budget} and~\ref{sec:reward}, respectively.
}

\begin{algorithm}[t]
\small
	\caption{Training method of \sexyname.}
	\label{alg:training}
	\begin{algorithmic}[1]\small
		\REQUIRE{
		Search space $\Omega$, latency distribution $\mB$, 
		learning rate $\eta$, training data set $\mD$, parameters $M$, $N$ and $K$.
		}
        \STATE Initialize model parameters $\theta$ for the generator and $w$ for the architecture evaluator. \\
        // \emph{Collect the architectures with accuracy and latency} \\
        \STATE Train a supernet $S$ on $\mD$. \\
        \STATE Randomly sample architectures $\left\{ \beta_i \right\}_{i=1}^{M}$ from $\Omega$. \\
        \STATE Construct tuples $\left\{( \beta_i, c(\beta_i), {\rm Acc}(\beta_i)) \right\}_{i=1}^{M}$ using $S$. \\
        // \emph{Learn the architecture evaluator} \\
        \WHILE{not convergent}
            \STATE Sample a set of latencies $\{B_k\}_{k=1}^{K}$ from $\mB$. \\
            \STATE Update the architecture evaluator by: \\
            \STATE ~~~~~~~~~$w \leftarrow w - \eta \nabla_w L(w)$. \\
            % \STATE ~~~~~$w \leftarrow w {-} \eta \frac{1}{KM(M{-}1)} \sum\limits_{k=1}^K \sum\limits_{i=1}^{M} \sum\limits_{j=1, j \neq i}^{M} \nabla_w \phi \big((R(\beta_i|T_k;w) {-} R(\beta_j|T_k;w) ) {\cdot} d( \beta_i, \beta_j, T_k) \big)$.
        \ENDWHILE \\
        // \emph{Learn the architecture generator}  \\
        \WHILE{not convergent}
            \STATE Sample a set of latencies $\{B_k\}_{k=1}^{K}$ from $\mB$. \\
            \STATE Obtain $\{\alpha_{\sss {B_k}}^{\sss (i)}\}_{i=1}^{N}$ from $\pi(\cdot|B_k; \theta)$ for each $B_k$. \\
            \STATE Update the generator via policy gradient by: \\
            \STATE ~~~~~~~~~$\theta \leftarrow \theta + \eta \nabla_\theta J(\theta)$. \\
            % \STATE ~~~~~$\theta \leftarrow \theta {+} \eta \frac{1}{KN}  \sum\limits_{k=1}^{K} \sum\limits_{i=1}^{N} \left[ \nabla_{\theta} \log \pi(\alpha_{\sss T_k}^{\sss (i)}|T_k;\theta) R(\alpha_{\sss T_k}^{\sss (i)}|T_k;w) {+} \lambda \nabla_{\theta} H(\pi(\cdot|T_k;\theta)) \right] $. \\
        \ENDWHILE
    \end{algorithmic}
\end{algorithm}

% It is worth noting that learning the Pareto frontier would benefit the {generation} performance of each scenario with a specific budget due to the shared knowledge across them (See results in Table~\ref{tab:mobile_comp}).
% To be specific, if we find a good architecture \wrt one budget, we can slightly change the width or depth of some modules to obtain promising architectures that satisfy the adjacent budgets.

\subsection{Vector Representation of Budget Bounds} \label{sec:rep_budget}
% Note that we consider a set of discrete budgets in training. 
% To alleviate the influence incurred by value difference of budgets,
% It is important to normalize them to make similar  
To represent different budgets,
following~\cite{pham2018efficient,guo2019nat}, we build a learnable embedding vector $\bb = g(B)$ for each sampled budget $B$. We incorporate these learnable embedding vectors into the parameters of the architecture generator and train them jointly. In this way, we are able to automatically learn the vectors of these budgets and encourage \sexyname to produce feasible architectures.  

As mentioned before, we only sample a set of discrete budgets to train \sexyname. To accommodate all the budgets belong to a continuous space, we propose an embedding interpolation method to represent a budget with any possible value.
Specifically, we perform a linear interpolation between the embedding of two adjacent discrete budgets to represent the considered budgets.
% Specifically, for 
For a target budget ${B}$ between two sampled budgets $B_1 {<} {B} {<} B_2$, the linear interpolation of the budget vector $\bb$ can be computed by
\begin{equation*}\label{eq:interpolation}
\begin{aligned}
  \bb = g({B}) = \xi  g(B_1) {+} (1 {-} \xi)  g(B_2),
  \text{~where~} \xi = \frac{{B_2} {-} B} {B_{2} {-} B_{1}},
\end{aligned}
\end{equation*}
Here, $\xi \in [0,1]$ denotes the weight of $B_1$ in interpolation.
% (See more details in Section~\ref{sec:train}).

\subsection{Learning the Architecture Evaluator $R(\cdot|B;w)$}\label{sec:reward}
% To obtain the reward from $R\left(\cdot| B; w \right)$, we introduce an architecture evaluator as $R\left(\cdot| B; w \right)$ and use the evaluator to estimate the performance of generated architectures under the budget $B$.
\qi{To guide the training of \sexyname, we shall propose a reward function to estimate the performance of architectures. Given any architecture $\beta$ and a budget $B$, we build an evaluator with three fully connected layers to predict the performance $R(\beta|B;w)$ of $\beta$ under the budget $B$. 
% The evaluator outputs a scalar as the reward for the scenario with $c(\beta) \leq B$.
Since we have no labeled data for training, we learn the evaluator via pairwise comparisons between any two architectures.
}

% In this section, we propose a Pareto dominance reward to train \sexyname.
% Specifically, {to learn the Pareto frontier, we need to find the Pareto optimal architecture \wrt each considered budget over multiple objectives (\eg, accuracy and latency).}
% In practice, one can approach the Pareto optimal solution by finding the Pareto improvement direction, which is a situation where some objectives will increase and no objectives will decrease. This situation is also called Pareto dominance where the better solution dominates the worse one.
% Thus, the key challenge to train \sexyname becomes how to judge whether an architecture dominates/outperforms another architecture under diverse budgets.

% \qi{
% \textbf{Architecture evaluator.} The Pareto dominance rule requires architecture pairs to find the Pareto optimal architectures. However, the generator model only finds an architecture at a time and thus we cannot be directly used to compute the reward.
% To address this issue, we propose to train an architecture evaluator $R(\cdot|T;w)$ to learn the proposed Pareto dominance rule
% and output a scalar as the reward for the scenario with $c(\alpha) \leq T$.
% }

\qi{\textbf{Pairwise ranking loss function.}
To obtain a promising evaluator, we train the architecture evaluator using a pairwise ranking loss, which has been widely used in ranking problems~\cite{freund2003efficient,burges2005learning,chen2009ranking}.
Specifically, we collect $M$ architectures with accuracy and latency, and record these architectures as a set of triplets $\{(\beta_i, c(\beta_i), {\rm Acc}(\beta_i))\}_{i=1}^{M}$.
Thus, given $M$ architectures, we have $M(M{-}1)$ architecture pairs $\{(\beta_{i}, \beta_{j})\}$ in total after omitting the pairs with the same architecture.
Assuming that we have $K$ budgets, the pairwise ranking loss becomes
\begin{equation}\label{eq:ranking_loss}
\begin{aligned}
    L(w) = &\frac{1}{KM(M{-}1)} \sum_{k=1}^K \sum_{i=1}^{M} \sum_{j=1, j \neq i}^{M} \phi \Big( d( \beta_{i}, \beta_{j}, B_k)    \\
     &\cdot \big[ R(\beta_{i}|B_k;w) - R(\beta_{j}|B_k;w) \big]  \Big),
\end{aligned}
\end{equation}
where $d\big(\beta_1, \beta_2, B_k \big)$ denotes a function to indicate whether $\beta_{i}$ is better than $\beta_{j}$ under the budget $B_k$. The $\phi(z) = \max (0, 1-z)$ is a hinge loss function.
We use the hinge loss $\phi(\cdot)$ to enforce the predicted ranking results $R(\beta_{i}|B_k;w) - R(\beta_{j}|B_k;w)$ to be consistent with the results of $d( \beta_{i}, \beta_{j}, B_k)$ obtained by a comparison rule based on Pareto dominance.
% Due to the page limit, we put more discussions of Eqn.~(\ref{eq:ranking_loss}) in the supplementary.
}

% \qi{To learn the evaluator $R(\beta|B;w)$, we collect $M$ architectures with accuracy and latency, and record these architectures as a set of triplets $\{(\beta_i, c(\beta_i), {\rm Acc}(\beta_i))\}_{i=1}^{M}$.}
% To learn the Pareto frontier, we propose an architecture evaluator to compute the reward based on the proposed Pareto dominance rule.
According to Eqn.~(\ref{eq:ranking_loss}), based on collected training data $\{(\beta_i, c(\beta_i), {\rm Acc}(\beta_i))\}_{i=1}^{M}$ and $\{B_k\}_{k=1}^{K}$, the gradient w.r.t. $w$ becomes
\begin{equation*}
\small
    \begin{aligned}
    \nabla_w L(w) {=} &\frac{1}{KM(M{-}1)} \sum_{k=1}^K \sum_{i=1}^{M} \sum_{j=1, j \neq i}^{M} \nabla_w \phi \Big( d( \beta_{i}, \beta_{j}, B_k)  \\
     &\cdot \big[ R(\beta_{i}|B_k;w) - R(\beta_{j}|B_k;w) \big]  \Big).
     \end{aligned}
\end{equation*}
% \qi{{Due to the page limit, we put more training details in the supplementary.}}

\subsection{Pareto Dominance Design}
\qi{ To compare the performance between two architectures, we need to define a reasonable function $d\big(\beta_1, \beta_2, B \big)$ in Eqn.~(\ref{eq:ranking_loss}). To this end, we define a Pareto dominance to guide the design of this function.
% we define a Pareto dominance rule to guide the learning of \sexyname. 
Specifically, Pareto dominance requires that the quality of an architecture should depend on both the satisfaction of budget and accuracy.
That means, given a specific budget $B$, a good architecture should be the one with the cost lower than or equal to $B$ and with high accuracy.
% Motivated by this, we use the Pareto dominance to compare two architectures and judge which one is dominative.
In this sense, we use Pareto dominance to compare two architectures and judge which one is dominative.
}

% In this sense, we can know which is better between any two architectures.
\kui{Given any two architectures $\beta_1, \beta_2$, if both of them satisfy the budget constraints (namely $c(\beta_1) \leq B$ and $c(\beta_2) \leq B$),  then $\beta_1$ dominates  $\beta_2$ if ${\rm Acc}(\beta_1) \geq {\rm Acc}(\beta_2)$. 
% Moreover,  if $c(\beta_1) \leq D < ~ c(\beta_2)$, clearly we have that $\beta_1$ dominates  $\beta_2$.
}
Moreover, when at least one of $\beta_1, \beta_2$ violates the budget constraint, clearly we have that $\beta_1$ dominates $\beta_2$ if $c(\beta_1) \leq c(\beta_2)$.
% Last, if both architectures violates the budget constraints, 
Formally, we define the Pareto dominance function $d\big(\beta_1, \beta_2, B \big)$ to reflect the above rules:
% \begin{footnotesize}
\begin{equation}\label{eq:compare_rule}
    d\big(\beta_1, \beta_2, B \big) = 
    \begin{cases} 
        ~1, ~~~~~{\rm if} ~~\left(c(\beta_1) \leq B ~{\land}~ c(\beta_2) \leq B \right) \\
        ~~~~~~~~~~~~~{\land}~ (\textcolor{blue}{{\rm Acc}(\beta_1) \geq {\rm Acc}(\beta_2)}); \vspace{3 pt} \\ 
               -1, ~~~{\rm else~if} ~~\left(c(\beta_1) \leq B ~{\land}~ c(\beta_2) \leq B \right) \\
        ~~~~~~~~~~~~~{\land}~ (\textcolor{blue}{{\rm Acc}(\beta_1) < {\rm Acc}(\beta_2)}); \vspace{3 pt} \\ 
        ~1, ~~~~~{\rm else~if} ~~c(\beta_1) \leq ~ c(\beta_2); \vspace{3 pt} \\ 
        -1, ~~~{\rm otherwise}.
    \end{cases}
\end{equation}
% \begin{equation}\label{eq:compare_rule} 
%     \begin{cases} 
%         \mathbbm{1}[{\rm Acc}(\beta_1) \geq {\rm Acc}(\beta_2)], ~{\rm if} ~~c(\beta_1) \leq T ~{\rm and}~ c(\beta_2) \leq T; \vspace{3 pt} \\ 
%         \mathbbm{1}[c(\beta_1) \leq c(\beta_2)], ~~~~~~~~~~~{\rm otherwise}.
%     \end{cases}
% \end{equation}
% \end{footnotesize}
Based on Eqn.~(\ref{eq:compare_rule}), we have $d(\beta_1, \beta_2, B) = - d(\beta_2, \beta_1, B)$ if $\beta_1 \neq \beta_2$. 

\begin{remark}
The accuracy constraint ${\rm Acc}(\beta_1) \geq {\rm Acc}(\beta_2)$ plays an important role in the proposed Pareto dominance function $d\big(\beta_1, \beta_2, B \big)$. Without the accuracy constraint, we may easily find the architectures with very low computation cost and poor performance (See results in Table~\ref{tab:diff_reward}.
\end{remark}

\begin{table*}[t!] 
    \caption{
    Comparisons with state-of-the-art architectures on mobile devices. $^*$ denotes the best architecture reported in the original paper. 
    % $^\dagger$ denotes the average search/generation results over 5 independent runs. 
    ``-'' denotes the results that are not reported. All the models are evaluated on $224 \times 224$ images of ImageNet. 
    }
    % \vspace{5pt}
    \centering
    \resizebox{1.0\textwidth}{!}
    {               
        \begin{tabular}{ccccccc}
        \toprule [0.15em]
        \multirow{2}[0]{*}{Architecture}  & \multirow{2}[0]{*}{Latency (ms)}  & \multicolumn{2}{c}{Test Accuracy (\%)} & \multirow{2}[0]{*}{\#Params (M)}  & \multirow{2}[0]{*}{\#MAdds (M)}   & Search Cost  \\
        \cline{3-4}
        & & Top-1 & Top-5 & & & (GPU Days) \\
        \midrule [0.1em]
        % MobileNetV3-Small (0.75$\times$)~\cite{howard2019searching} & 30.9 & 65.4 & - & 2.0 & 44 \\
        % MobileNetV3-Small (1.0$\times$)~\cite{howard2019searching} & 39.8 & 67.4 & - & 2.4 & 56  \\
        MobileNetV3-Large (0.75$\times$)~\cite{howard2019searching} & 93.0 & 73.3 & - & 4.0 & 155 & -  \\
        MobileNetV2 (1.0$\times$)~\cite{sandler2018mobilenetv2} & 90.3 & 72.0 & - & 3.4 & 300 & - \\
        FBNetV2~\cite{wan2020fbnetv2} & - & 76.3 & 92.9 & - & 321 & 30.0 \\
        % MnasNet-A1 (0.5$\times$)~\cite{tan2019mnasnet} & 37.51 & 68.9 & 88.4 & 2.1 & 105 \\
        % OFA-S-80 & 76.8 & 76.8 & 93.3 & 6.1	& 350 \\
        OFA-80~\cite{cai2019once} & 76.8 & 76.8 & 93.3 & 6.1	& 350 & 51.7 \\
        % OFA-MO-80 & 77.6 & 76.6 & 93.2 & 7.9 & 340 & 51.7 \\
        % OFA-100~\cite{cai2019once} & 99.7 & 76.8 & 93.4 & 6.1 &	350 \\
        % \sexyname-80 & 79.9 & \textbf{77.5} & \textbf{93.7} & 7.3 & 349 & 0.7 \\
        \sexyname-80 & 79.9 & \textbf{78.4} & \textbf{94.0}  & 7.3 & 349 & 0.7 \\
        % \sexyname-100 & 93.9 & \textbf{78.2} & \textbf{94.0} & 9.1 & 405 \\
        \midrule
        FBNet-A~\cite{wu2019fbnet} & 91.7 & 73.0 & - & 4.3 & 249 & 9.0 \\
        ProxylessNAS-Mobile~\cite{cai2018proxylessnas} & 97.3 & 74.6 & - & 4.1 & 319 & 8.3 \\
        ProxylessNAS-CPU~\cite{cai2018proxylessnas} & 98.5 & 75.3 & - & 4.4 & 438 & 8.3 \\
        % MnasNet-A1 (0.75$\times$)~\cite{tan2019mnasnet} & 102.9 & 73.3 & 91.3 & 2.9 & 227  \\
        MobileNetV3-Large (1.0$\times$)~\cite{howard2019searching} & 107.7 & 75.2 & - & 5.4 & 219 & - \\
        % OFA-S-110 & 109.2 & 77.5 & 93.6 & 6.4 & 406 \\
        OFA-110~\cite{cai2019once} & 109.3 & 78.1 & 94.0 & 10.2 & 482 & 51.7 \\
        % OFA-MO-110 & 106.3 & 78.0 & 93.8 & 8.4 & 478 & 51.7 \\
        % OFA-125~\cite{cai2019once} & 124.9 & 77.7 &	93.7 & 6.4 & 406 \\
        % \sexyname-110 & 106.8 & \textbf{78.4} & \textbf{94.2} & 9.9 & 451 & 0.7 \\
        \sexyname-110 & 106.8 & \textbf{79.5} & \textbf{94.5} & 9.9 & 451 & 0.7 \\
        % \sexyname-125 & 124.5 & 78.6 & 94.2 & 8.7 & 459 \\
        \midrule
        RandWire~\cite{xie2019exploring} & - & 74.7 & 92.2 & 5.6 & 583 & - \\
        ProxylessNAS-GPU~\cite{cai2018proxylessnas} & 123.3 & 75.1 & - & 7.1 & 463 & 8.3  \\
        MobileNetV2 (1.4$\times$)~\cite{sandler2018mobilenetv2} & 139.6 & 74.7 & - & 6.9 & 585 & - \\
        MnasNet-A1 (1.0$\times$)~\cite{tan2019mnasnet} & 120.7  & 75.2 & 92.5 & 3.4 & 300 & $\sim$3792  \\
        % FBNet-B~\cite{wu2019fbnet} & 113.4 & 74.1 & - & 4.5 & 295  \\
        FBNet-C~\cite{wu2019fbnet} & 135.2 & 74.9 & - & 5.5 & 375 & 9.0 \\
        % OFA~\cite{cai2019once} & 115.4 & 78.4 & 94.1 & 8.4 & 388 \\
        % OFA-S-140 & 130.0 & 77.7 & 93.7 & 6.6 & 428 \\
        OFA-140~\cite{cai2019once} & 133.7 & 78.4 & 94.1 & 9.1 & 488 & 51.7 \\
        % OFA-MO-140 & 139.0 & 78.4 & 94.0 & 9.5 & 486 & 51.7 \\
        % OFA-150~\cite{cai2019once} & 150.0 & 77.6 &	93.7 & 6.6 & 428 \\
        % \sexyname-140 & 127.8 & \textbf{78.8} & \textbf{94.3}  & 9.2 & 492 & 0.7 \\
        \sexyname-140 & 127.8 & \textbf{79.8} & \textbf{94.7}  & 9.2 & 492 & 0.7 \\
        % \sexyname-150 & 144.5 & 78.8 & 94.4 & 10.2 & 533 \\
        \midrule
        NSGANetV1~\cite{lu2020multi} & - & 76.2 & 93.0 & 5.0 & 585 & 27 \\
        PONAS-C~\cite{huang2020ponas} & 145.1 & 75.2 & - & 5.6 & 376 & 8.8 \\
        DARTS~\cite{liu2018darts} & 176.6 & 73.1 & 91.0 &  4.7 & 574 & 4.0 \\
        P-DARTS~\cite{chen2019progressive} & 168.7 & 75.6 & 92.6 & 4.9 & 577 & 3.8 \\ 
        % OFA-S-170 & 163.6 & 78.2 & 94.1 & 7.8 & 534 \\
        OFA-170~\cite{cai2019once} & 168.3 & 78.9 & 94.4 & 10.7 & 661 & 51.7 \\
        % OFA-MO-170 & 165.0 & 78.8 & 94.4 & 8.5 & 584 & 51.7 \\
        % OFA-175~\cite{cai2019once} & 174.8 & 78.2 &	94.1 & 7.8 & 534 \\
        % \sexyname-170 & 167.1 & \textbf{79.0} & \textbf{94.5} & 10.0 & 606 & 0.7 \\
        \sexyname-170 & 167.1 & \textbf{80.3} & \textbf{95.1} & 10.0 & 606 & 0.7 \\
        % \sexyname-175 & 171.0 & 78.9 & 94.5 & 9.8 & 577 \\
        \midrule
        NSGANetV2~\cite{lu2020nsganetv2} & - & 79.1 & 94.5 & 8.0 & 400 & 1 \\
        NAGO~\cite{ru2020neural} & - & 76.8 & 93.4 & 5.7 & - & 20.0 \\
        ENAS~\cite{pham2018efficient} & 232.4 & 73.8 & 91.7 & 5.6 & 607 & 0.5 \\
        PC-DARTS~\cite{xu2020pcdarts} & 194.1 & 75.8 & 92.7 & 5.3 & 597 & 0.1 \\
        MnasNet-A1 (1.4$\times$)~\cite{tan2019mnasnet} & 205.5 & 77.2 & 93.5 & 6.1 & 592 & $\sim$3792 \\
        EfficientNet B0~\cite{EfficientNet} & 237.7 & 77.3 & 93.5 & 5.3 & 390 & - \\
        % OFA-S-200 & 197.5 & 78.3  & 94.2 & 8.4 &	629 \\
        Cream-L~\cite{peng2020cream} & - & 80.0 & 94.7 & 9.7 & 604 & 12 \\
        OFA$^*$~\cite{cai2019once} & 201.9 & 80.2 & 95.1 & 9.1 & 743 & 51.7 \\
        OFA-200~\cite{cai2019once} & 195.9 & 79.0  & 94.5 & 11.0 & 783 & 51.7 \\
        % OFA-MO-200 & 187.4 & 78.9  & 94.4 & 9.1 & 630 & 51.7 \\
        % \sexyname-200  & 193.9 & \textbf{79.2} & \textbf{94.7} & 10.4 & 724 & 0.7 \\
        \sexyname-200 & 193.9 & \textbf{80.5} & \textbf{95.3}  & 10.4 & 724 & 0.7 \\
        \bottomrule[0.15em]
        \end{tabular}  
    }
    \label{tab:mobile_comp}     
\end{table*}

% \textbf{Training the architecture generator $f(T;\theta)$.}

\begin{figure*}[t]
	\centering
% 	\vspace{-15pt}
	\subfigure[Ground-truth latency histogram.]{
		\includegraphics[width = 0.68\columnwidth]{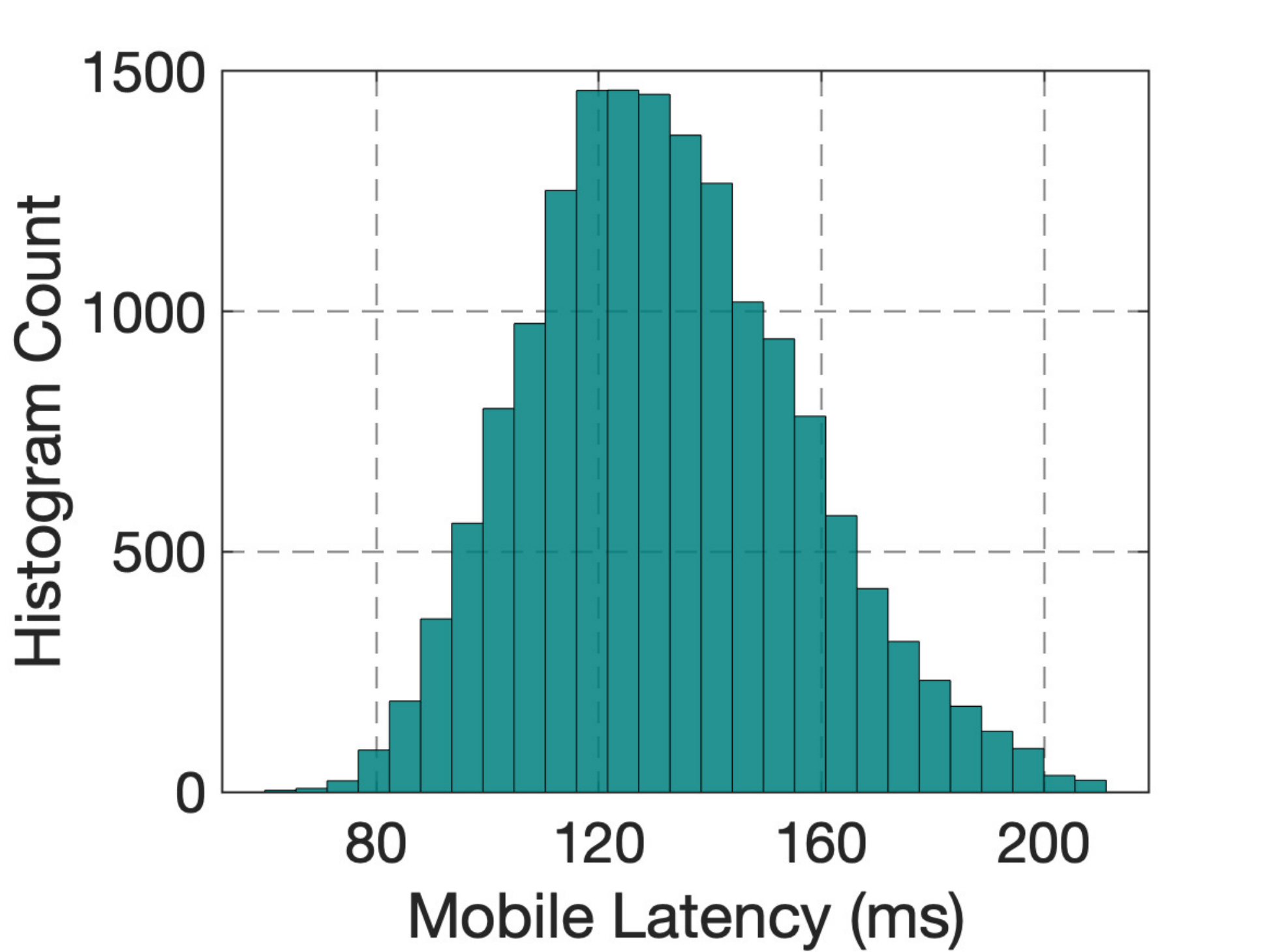}\label{fig:mobile_dist}
	}~~
	\subfigure[{Generation results with $B{=}110$ms.}]{
		\includegraphics[width = 0.68\columnwidth]{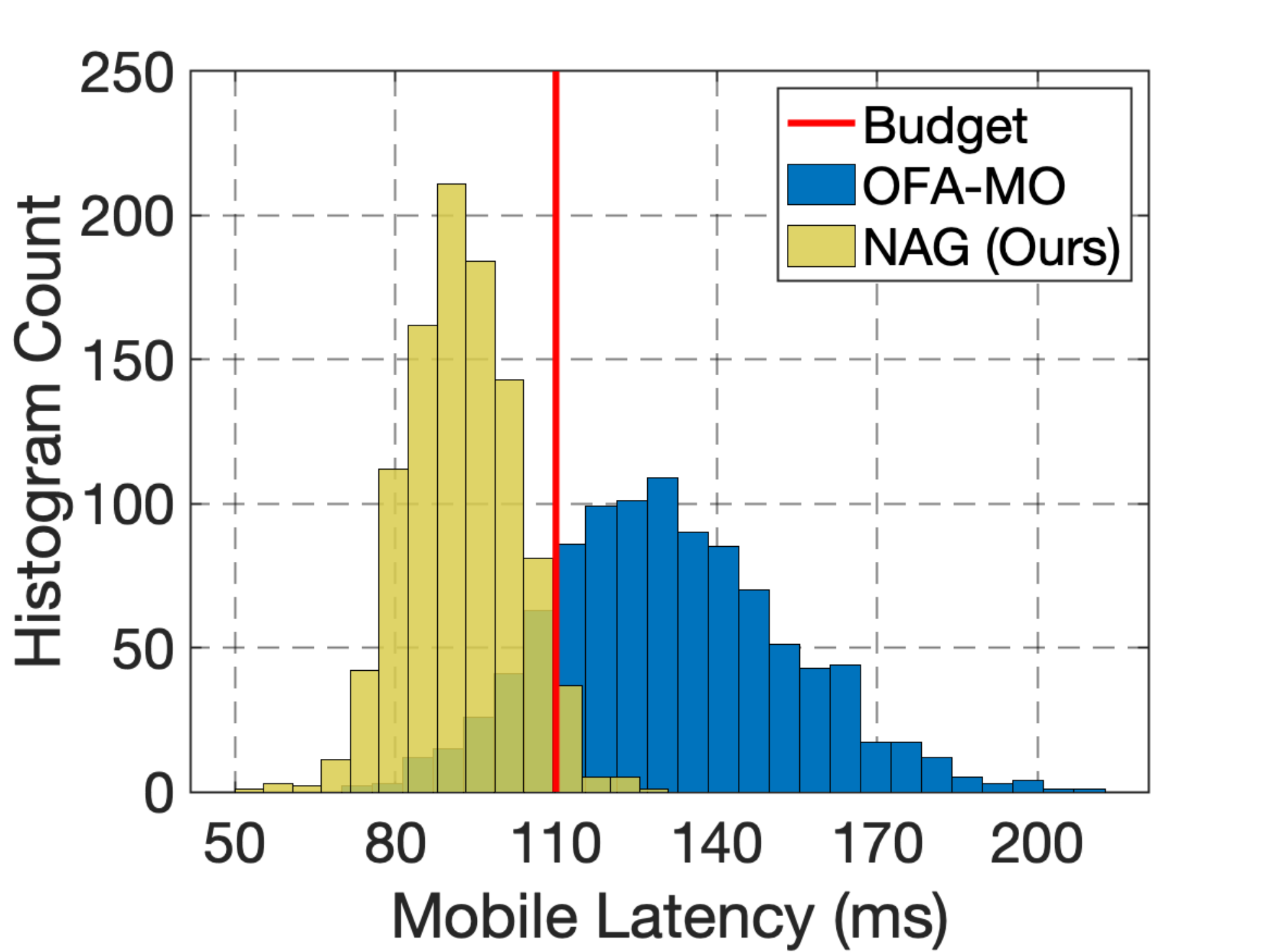}\label{fig:histogram_110}
	}~~
	\subfigure[Generation results with $B{=}140$ms.]{
		\includegraphics[width = 0.68\columnwidth]{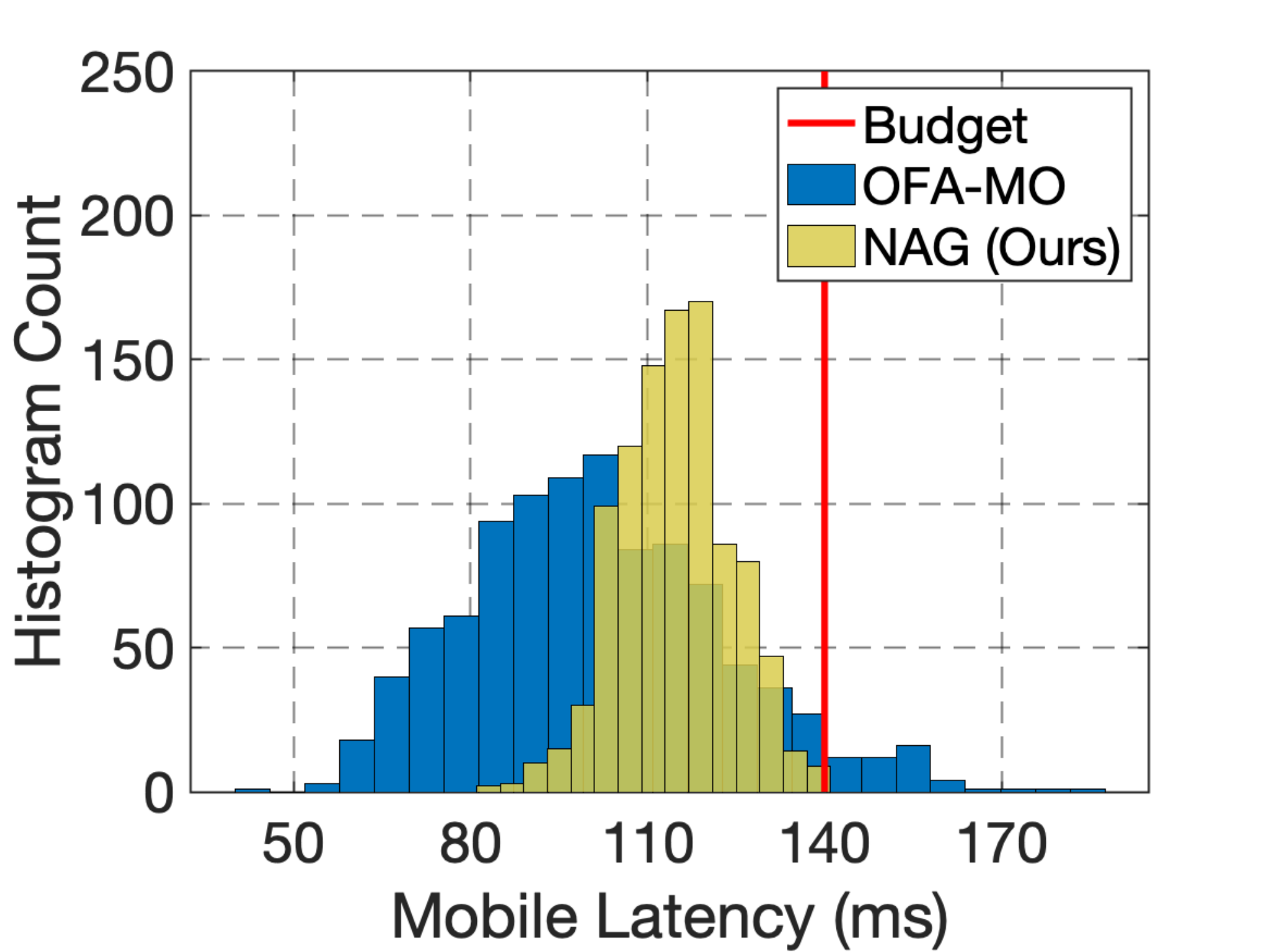}\label{fig:histogram_140}
	}
% 	\vspace{10pt}
	\caption{
    % Latency histograms and Pareto curves of the architectures on mobile devices. 
    Latency histograms {of sampled architectures} on mobile devices. (a) Ground-truth latency histogram of $16,000$ architectures that are uniformly sampled from the search space. (b) The latency histogram of $1,000$ architectures sampled by different methods given $B{=}110$ms. {(c) The latency histogram of $1,000$ architectures sampled by different methods given $B{=}140$ms.}}
	\label{fig:distribution}
% 	\vspace{-10pt}
\end{figure*}

\subsection{Inferring Architecture for Required Budgets}

During inference, we allow the input budget to be any value. Based on the set of discrete budgets in training, we first use linear interpolation to obtain the desired budget, and then perform inference using \sexyname to produce architectures.
% based on the learned policy $\pi(\cdot|T; \theta)$, we conduct sampling to find promising architectures.
Specifically, given an arbitrary budget $B$, we first sample several candidate architectures from the learned policy $\pi(\cdot|B; \theta)$. 
If all the sampled architectures violate the budget constraint, we will repeat the sampling process until finding a feasible architecture.
Then, we select the feasible architecture with the highest validation accuracy.

% Note that we train \sexyname using a finite number of discrete budgets. 
% !!!
% To enable $T$ to be any value in a continuous space, we perform a linear interpolation between the embedding of two adjacent discrete budgets to represent the considered budgets.
% Specifically, for a target budget $\hat{T}$ between two sampled budgets $T_1 {<} \hat{T} {<} T_2$, the linear interpolation of the budget embedding $g(\hat{T})$ is formulated as
% \begin{equation}\label{eq:interpolation}
% \begin{aligned}
%   g(\hat{T}) := &\lambda \cdot g(T_1) + (1 - \lambda) \cdot g(T_2), \\
%   \text{where~} &\lambda = \frac{\hat{T} - T_{1}}{T_{2} - T_{1}}.
% \end{aligned}
% \end{equation}

% \textbf{Training cost of \sexyname.} The training of {the} super network takes about 0.6 GPU days. 
% % The data preparation process takes about 40 GPU hours. 
% We train the architecture evaluator for about 15 minutes. The training of the generator takes about 15 minutes, which is more efficient than the search process of most NAS methods. 

\begin{figure}[t!]
	\centering
	\includegraphics[width = 0.90\columnwidth]{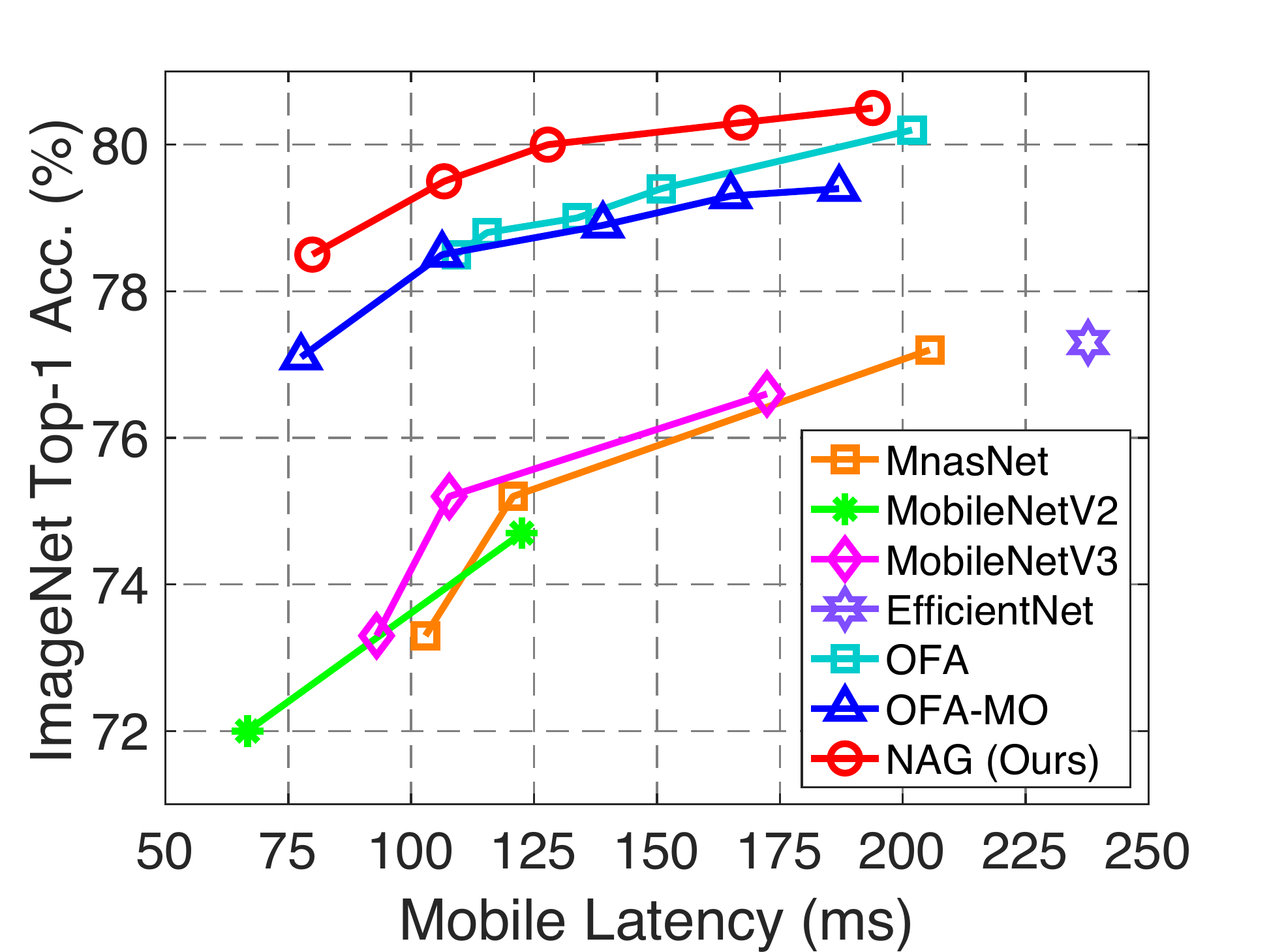}
% 	\subfigure[Results on Intel Core i5 CPU.]{
% 		\includegraphics[width = 0.70\columnwidth]{./cpu_compare.pdf}\label{fig:cpu_compare}
% 	}~
% 	\subfigure[Results on TITAN X GPU.]{
% 		\includegraphics[width = 0.70\columnwidth]{./gpu_compare.pdf}\label{fig:gpu_compare}
% 	}
	\caption{Comparisons of the architectures obtained by different methods on mobile devices.}
	\label{fig:result_compare}
% 	\vspace{5pt}
\end{figure}

\section{Experiments}\label{sec:exp}

We apply \sexyname to produce architectures under diverse latency budgets on three kinds of hardware platforms, including mobile devices (equipped with a Qualcomm Snapdragon 821 processor), CPU devices (Intel Core i5-7400), and GPU devices (NVIDIA TITAN X).
For convenience, we use ``Architecture-$B$'' to represent the {generated} architecture that satisfies the latency budget w.r.t. $B$, \eg, \sexyname-80.
Due to the page limit, we put the experimental results on CPU and GPU devices and visualizations of the {generated} architectures in the supplementary.

\textbf{Implementation details.}
We use MobileNetV3~\cite{howard2019searching} as the backbone to build the search space.
Following~\cite{wu2019fbnet}, we randomly choose 10\% classes from ImageNet as the training set 
% to train a supernet.
% We 
and train the supernet with progressive shrinking strategy~\cite{cai2019once} for 90 epochs.
To obtain the range of latency, we randomly sample $M{=}16,000$ architectures from the search space (See Figure~\ref{fig:mobile_dist}).
Then, we evenly sample $K{=}10$ discrete budgets to divide the range of latency (See the discussions on $K$ in Section~\ref{sec:effect_k}). 
Note that we share the same search space with OFA~\cite{cai2019once}.
We set the dimension of the embedding vector of budgets to 64.
To accelerate model evaluation, following~\cite{cai2019once,lu2020neural}, we first obtain the parameters from \lhk{the} full network of OFA and then finetune them for 75 epochs to obtain the final performance. 
% Due to the page limit, 
We put more details in the supplementary.

\subsection{Comparisons with State-of-the-art Methods}
We compare \sexyname with state-of-the-art methods on mobile device. We also consider the following baselines:
1) \textbf{OFA} uses the evolutionary search method~\cite{real2019regularized} used in the OFA paper~\cite{cai2019once} to perform architecture search.
2) \textbf{OFA-MO} conducts architecture search based on OFA super network by exploiting the multi-objective reward~\cite{tan2019mnasnet}.
% 3) \textbf{OFA-S} finds the best one from the $16,000$ architectures sampled from the learned super network.
From Table~\ref{tab:mobile_comp} and Figure~\ref{fig:result_compare}, our \sexyname consistently achieves higher accuracy than other methods. 
{Compared with the methods that find architectures for each budget independently
(\eg, OFA and OFA-MO), our \sexyname only needs to search once to generate the architectures with arbitrary target latency.}

We also visualize the latency histograms of the architectures generated on mobile devices in Figure~\ref{fig:histogram_110} and {Figure~\ref{fig:histogram_140}}.
{Given latency budgets of $110$ms and $140$ms, OFA-MO is prone to produce a large \lhk{number} of architectures that cannot satisfy the target budgets. 
These results show that it is hard to design the multi-objective reward to obtain the preferred architectures.}
Instead, \sexyname uses the Pareto dominance reward to encourage the architectures to satisfy the desired budget constraints. 
In this sense, most architectures generated by our \sexyname are able to fulfill the target budgets.
We put more visual results of latency histograms with other target latencies in the supplementary.

\begin{figure}[t!]
    % 	\subfigure[Comparisons of Pareto frontiers.]{
    % 		\includegraphics[width = 0.9\columnwidth]{./pareto_curve.pdf}\label{fig:pareto_curve}
    % 	}
    \centering
    \includegraphics[width = 0.90\columnwidth]{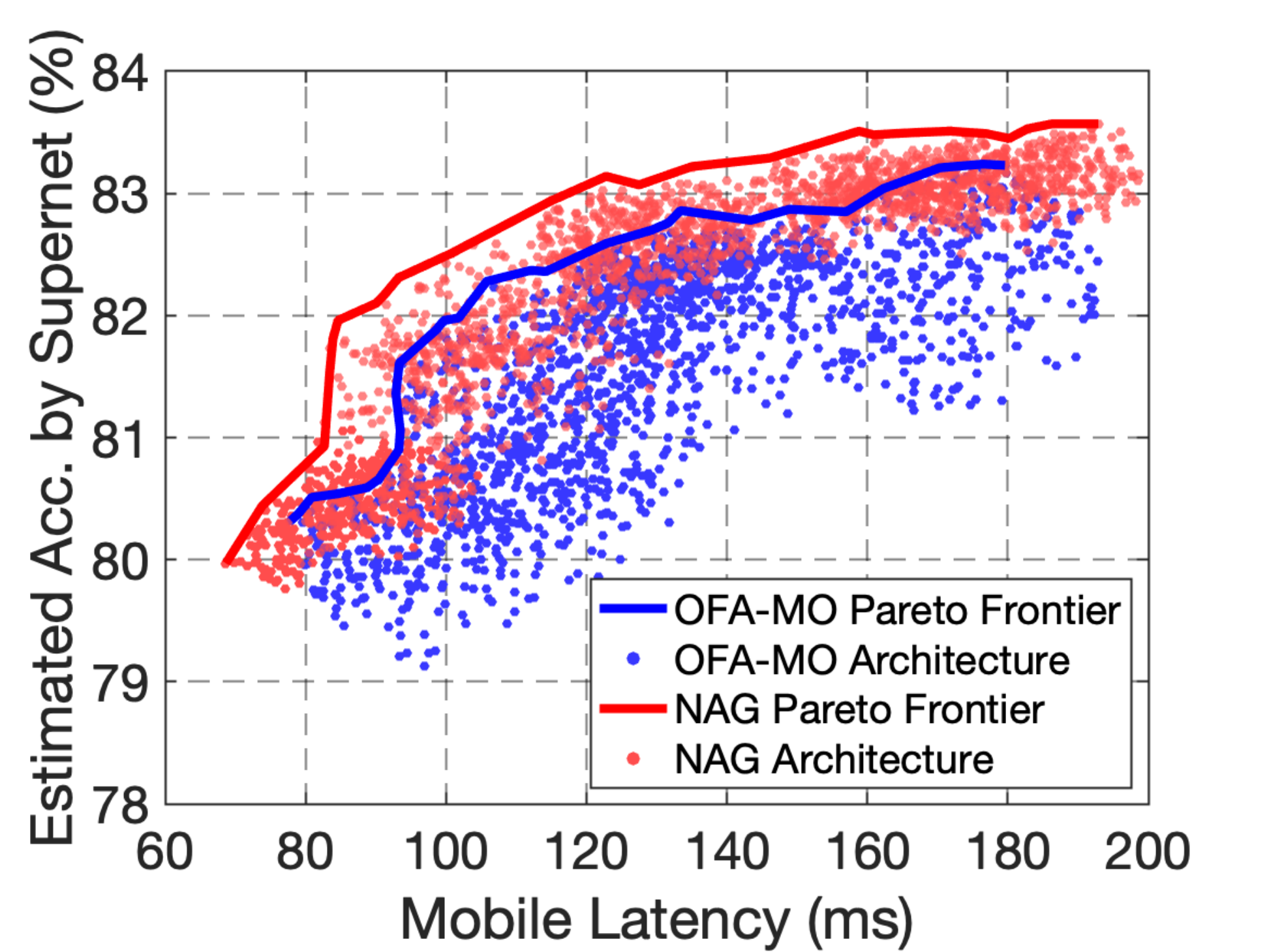}
	\caption{Comparisons of the Pareto frontiers of the {generated} architectures using different methods.}
	\label{fig:pareto_curve}
\end{figure}

\begin{table*}[t!]
  \centering
  \caption{Comparisons of different reward functions based on \sexyname. We report the latency on mobile devices.
%   For convenience, we use Acc. and Lat. to represent accuracy on ImageNet and latency on mobile device, respectively. 
%   $P_{s} (\%)$ denotes the proportion of the {generated} architectures that satisfy the corresponding budget.
  }
    \resizebox{1.0\textwidth}{!}
    {      
    \begin{tabular}{c|cc|cc|cc|cc|cc}
    \toprule
    \multirow{2}[0]{*}{Reward}  & \multicolumn{2}{c|}{$B_1 {=} 80$ms} & \multicolumn{2}{c|}{$B_2 {=} 110$ms} & \multicolumn{2}{c|}{$B_3 {=} 140$ms} & \multicolumn{2}{c|}{$B_4 {=} 170$ms} & \multicolumn{2}{c}{$B_5 {=} 200$ms} \\
    % \cline{3-12}
      & \multicolumn{1}{c}{Acc. (\%)} & \multicolumn{1}{c|}{Lat. (ms)} & \multicolumn{1}{c}{Acc.} & \multicolumn{1}{c|}{Lat. (ms)}  & \multicolumn{1}{c}{Acc.} & \multicolumn{1}{c|}{Lat. (ms)} & \multicolumn{1}{l}{Acc.} & \multicolumn{1}{c|}{Lat. (ms)} & \multicolumn{1}{l}{Acc.} & \multicolumn{1}{l}{Lat. (ms)} \\
    \hline
    % \multirow{2}[0]{*}{Multi-objective Reward~\cite{tan2019mnasnet}}  & 76.6    &  1.4     &    78.0   &  33.2     &   78.4    &    54.0   &   78.8    &    90.5   &   78.9    &    99.9    \\
    \multirow{1}[0]{*}{Multi-objective Reward~\cite{tan2019mnasnet}} & 77.0   &  77.6  &   78.5  &  106.3    &   78.9  & 139.0  &   79.3   &  165.1  &  79.5    &   187.3  \\
    % \multirow{2}[0]{*}{Multi-objective Absolute Reward~\cite{Bender2020TuNAS}}  & 76.1    &  5.7     &    77.4   &  37.6     &   78.1    &    51.8   &   78.7    &    87.2   &   78.7    &    99.6    \\
    \multirow{1}[0]{*}{Multi-objective Absolute Reward~\cite{Bender2020TuNAS}} & 78.1   &  76.8   &   78.9  &  109.2 &   79.2   &  130.1    &   79.5   & 163.6  &  79.9  & 197.5 \\
    % \multirow{2}[0]{*}{Pareto Dominance Reward (Ours)}   & 76.3 & 92.8 & 78.1 & 92.5 & 78.6 & 93.2 & 78.9 & 94.3 & 79.0 & 99.5\\
    % \multirow{1}[0]{*}{Pareto Dominance Reward (Ours)} &    \textbf{77.2}      &  \textbf{78.4}    &   \textbf{78.7}  &   \textbf{79.0}      &   \textbf{79.2}   \\
    \hline
    \multirow{1}[0]{*}{Pareto Dominance Reward (w/o accuracy constraint)} &   73.8      &  74.4    &  73.6   &    64.9     &  74.3  &  66.5 & 73.9  &  70.0  & 74.0  &  70.8  \\
    \multirow{1}[0]{*}{Pareto Dominance Reward (Ours)} &    \textbf{78.4}  & 79.9    &  \textbf{79.5}  & 106.8  &   \textbf{79.8}  & 127.8 &  \textbf{80.3}  & 167.1    &   \textbf{80.5} & 193.9  \\
    
    \bottomrule
    \end{tabular}%
    }
  \label{tab:diff_reward}%
\end{table*}%

\begin{table*}[t!]
  \centering
  \caption{Effect of different search strategies on the performance of \sexyname. We report the accuracy on ImageNet. 
%   $P_{s} (\%)$ denotes the proportion of the {generated} architectures that satisfy the corresponding budget.
  }
    \resizebox{0.70\textwidth}{!}
    {      
    \begin{tabular}{c|c|c|c|c|c}
    \toprule
    Search Strategy  & \multicolumn{1}{c|}{$B_1 {=} 80$ms} & \multicolumn{1}{c|}{$B_2 {=} 110$ms} & \multicolumn{1}{c|}{$B_3 {=} 140$ms} & \multicolumn{1}{c|}{$B_4 {=} 170$ms} & \multicolumn{1}{c}{$B_5 {=} 200$ms} \\
    % \cline{3-12}
    %   & \multicolumn{1}{c}{Acc.} & \multicolumn{1}{c}{Acc.} & \multicolumn{1}{c}{Acc.} & \multicolumn{1}{l}{Acc.} & \multicolumn{1}{l}{Acc.}  \\
    \hline
    Repeated Independent Search  & 76.7 &   78.6       &   79.1     &   79.4       &  79.7         \\
    Pareto Frontier Search &    \textbf{78.4}      &  \textbf{79.5}    &   \textbf{79.8}  &   \textbf{80.3}      &   \textbf{80.5}   \\
    
    \bottomrule
    \end{tabular}%
    }
  \label{tab:pareto_learning}%
\end{table*}%

Moreover, we compare the learned/searched frontiers of different methods
{and show the comparisons of Pareto frontiers in Figure~\ref{fig:pareto_curve}.
We plot all the architectures produced by different methods to form the Pareto frontier. 
Specifically, we use the architectures searched by multiple independent runs under different budgets for OFA-MO.
For \sexyname, we use linear interpolation to generate architectures that satisfy different budgets.}
From Figure~\ref{fig:pareto_curve}, our \sexyname finds a better frontier than OFA-MO due to the shared knowledge across the search process under different budgets.

% We also evaluate \sexyname on Intel Core i5-7400 CPU and NVIDIA TITAN X GPU. 
% From Figure~\ref{fig:result_compare}, 
% \sexyname consistently find{s} better architectures than existing methods for each latency budget on all considered devices (See more results in {the} supplementary). 

\subsection{Ablation Studies}\label{sec:ablation}

% In this experiment, 
Here, we investigate the effectiveness of the Pareto frontier learning strategy and the Pareto dominance reward.
From Table~\ref{tab:diff_reward},
the Pareto frontier learning strategy tends to find better {architectures} than the independent search process due to the shared knowledge across the search processes under different budgets. 
Compared with two existing multi-objective rewards~\cite{tan2019mnasnet,Bender2020TuNAS},
% multi-objective reward {and multi-objective absolute reward}, 
the Pareto dominance reward encourages the generator to produce architectures that satisfy the considered budget constraints.
% For example, even if only a few architectures have the latency lower than $80$ms (See Figure~\ref{fig:mobile_dist}), with the Pareto dominance reward, there are still $92.8\%$ architectures generated by \sexyname satisfying the target budget.
\qi{Moreover, if we do not consider accuracy constraint in the Pareto dominance reward, the generated architectures have low latency and poor accuracy.}
With both the Pareto frontier learning strategy and the Pareto dominance reward, our \lhk{method} yields the best results under all budgets.

\begin{table}[t]
    \centering
    \caption{
    Comparisons of the time cost for architecture generation/design among different methods.}
    \resizebox{0.45\textwidth}{!}
    {      
    \begin{tabular}{c|cccc} 
    \toprule
    Method                &  \sexyname  & PC-DARTS   & ENAS  & DARTS  \\ 
    \midrule
    Time Cost &   {$\leq$5 s}   & 2 hours     & 12 hours & 4 days  \\
    % \sexyname            &   2h   & 2h      & 2h     & 2h    & 2h  \\
    \bottomrule
    \end{tabular}
    }
    \label{tab:generation_cost}
\end{table}

\begin{table}[t]
  \centering
  \caption{Effect of $K$ on the generation performance of \sexyname. 
  We compare the {generated} architectures using different values of $K$ with the target latency $B{=}140$ms on ImageNet.}
  \resizebox{0.36\textwidth}{!}
   {
    \begin{tabular}{c|ccccc}
    \toprule
    $K$      &  2     & 5     & 10    & 30  \\
    \midrule
    Top-1 Acc. (\%) & 79.1 & 79.4 & \textbf{79.8} & 79.7  \\
    % Top-1 Acc. (\%) & 78.1 & 78.7 & \textbf{78.8} & 78.7  \\
    % $\overline{P_{s}}$ (\%)   &   89.1    &  75.4     &   73.4    & 70.7  & 66.8 \\
    % $\overline{P_{s}}$ (\%)   &   96.3    &   86.0    &   90.8    & 88.0 & \\
    \bottomrule
    \end{tabular}%
    }
  \label{tab:effect_k}%
\end{table}%

\subsection{Comparisons of Architecture Generation Cost}
{
In this part, we compare the architecture generation cost of different methods for 5 different budgets and show the comparison results in Table~\ref{tab:generation_cost}.
Given an arbitrary target budget, existing NAS methods need to perform an independent search to find feasible architectures. 
By contrast, since \sexyname directly learns the whole Pareto frontier, we are able to generate promising architectures based on a learned generator model via inference.
Thus, the architecture generation cost of \sexyname is much less than other existing methods (See results in Table~\ref{tab:generation_cost}). In this sense, we are able to greatly accelerate the architecture design process in real-world scenarios.
These results demonstrate the efficiency of our \sexyname in generating architectures.
}

\subsection{Effect of $K$ on {the Generation Performance}}\label{sec:effect_k}

In this part, we investigate the effect of $K$ on the generation performance of \sexyname based on mobile device.
Note that we evenly select $K$ budgets from the range of latency. 
To investigate the effect of $K$, we consider several candidate values of {$K \in \{ 2, 5, 10, 30 \}$.}
{
We show the Top-1 accurac{ies} of the architectures generated by \sexyname with different $K$ on ImageNet in Table~\ref{tab:effect_k}.
Since a small number of selected budgets $K$ cannot accurately approximate the ground-truth Pareto frontier or provide enough shared knowledge between different search processes, our method yields poor results with $K=2$.
When we increase $K$ larger than 5, we are able to greatly improve the performance of the generated architectures. From Table~\ref{tab:effect_k}, our method yields the best result with $K=10$.
Therefore, we select $K=10$ discrete budgets to train our \sexyname model in the experiments.}

\section{Conclusion}
In this paper, we have proposed a novel Pareto-Frontier-aware Neural Architecture {Generator} (\sexyname) method to produce effective architectures for any given budget.
Specifically, we train the \sexyname model to learn the Pareto frontier by maximizing the expected reward over a set of budgets.
{Moreover, we propose an embedding interpolation method to adapt the learned generator model to any possible budget.}
{
To provide accurate rewards to guide the learning of \sexyname, we propose a Pareto dominance reward that is able to judge whether an architecture is better than another.
}
Based on the learned Pareto frontier, \sexyname is able to produce promising architectures under diverse budgets. 
Extensive experiments on three platforms (\ie, mobile, CPU, and GPU devices) demonstrate the effectiveness of the proposed method.

\clearpage

% In the unusual situation where you want a paper to appear in the
% references without citing it in the main text, use \nocite
\nocite{langley00}

\bibliography{example_paper}
\bibliographystyle{icml2021}

\end{document}